\documentclass[11pt,a4paper]{article}
\usepackage[hyperref]{emnlp2020}
\usepackage{times}
\usepackage{latexsym}

\usepackage{subcaption}
\usepackage{graphicx}
\usepackage{booktabs}
\usepackage{multirow}
\usepackage{makecell}
\usepackage{xspace}
\usepackage{url}

\newcommand{\TransQuest}{TransQuest\xspace}

\usepackage{microtype}

\aclfinalcopy 

\title{TransQuest at WMT2020: Sentence-Level Direct Assessment}

\author{Tharindu Ranasinghe$^\diamondsuit$, \textbf{Constantin Or\u{a}san$^\heartsuit$ and Ruslan Mitkov$^\diamondsuit$} \\
 $^\diamondsuit$Research Group in Computational Linguistics, University of Wolverhampton, UK \\
 $^\heartsuit$Centre for Translation Studies, University of Surrey, UK \\
 {\tt \{t.d.ranasinghehettiarachchige, r.mitkov\}@wlv.ac.uk} \\
 {\tt  c.orasan@surrey.ac.uk} }

\date{}

\begin{document}

\maketitle
\begin{abstract}
This paper presents the team \TransQuest's participation in Sentence-Level Direct Assessment shared task in WMT 2020. We introduce a simple QE framework based on cross-lingual transformers, and we use it to implement and evaluate two different neural architectures. The proposed methods achieve state-of-the-art results surpassing the results obtained by OpenKiwi, the baseline used in the shared task. We further fine tune the QE framework by performing ensemble and data augmentation. Our approach is the winning solution in all of the language pairs according to the WMT 2020 official results.  
\end{abstract}

\section{Introduction}

The goal of quality estimation (QE) systems is to determine the quality of a translation without having access to a reference translation. This makes it very useful in translation workflows where it can be used to determine whether an automatically translated sentence is good enough to be used for a given purpose, or if it needs to be shown to a human translator for translation from scratch or postediting \cite{kepler-etal-2019-openkiwi}. Quality estimation can be done at different levels: document level, sentence level and word level \cite{ive-etal-2018-deepquest}. This paper presents \TransQuest, a sentence-level quality estimation framework which is the winning solution in all the language pairs in the WMT 2020 Sentence-Level Direct Assessment shared task \cite{specia2020findings}.

In the past, high preforming quality estimation systems such as QuEst \cite{specia-etal-2013-quest} and QuEst++ \cite{specia-etal-2015-multi} were heavily dependent on linguistic processing and feature engineering. These features were fed into traditional machine-learning algorithms like support vector regression and randomised decision trees \cite{specia-etal-2013-quest}, which then determined the quality of a translation. Even though, these approaches provide good results, they are no longer the state of the art, being replaced in recent years by neural-based QE systems which usually rely on little or no linguistic processing. For example the best-performing system at the WMT 2017 shared task on QE was \textsc{POSTECH}, which is purely neural and does not rely on feature engineering at all \cite{kim-etal-2017-predictor}. 

In order to achieve high results, approaches such as \textsc{POSTECH} require extensive pre-training, which means they depend on large parallel data and are computationally intensive \cite{ive-etal-2018-deepquest}. \TransQuest, our QE framework removes this dependency on large parallel data by using crosslingual embeddings \cite{transquest:2020} that are already fine-tuned to reflect properties between languages \cite{10.1613/jair.1.11640}. \citet{transquest:2020} show that by using them, TransQuest eases the burden of having complex neural network architectures, which in turn entails a reduction of the computational resources. That paper also shows that TransQuest performs well in  transfer learning settings where it can be trained on language pairs for which we have resources and applied successfully on less resourced language pairs. 

The remainder of the paper is structured as follows. The dataset used in the competition is briefly discussed in Section \ref{sec:datasets}. In Section \ref{sec:methodology} we present the \TransQuest framework and the methodology employed to train it. This is followed by the evaluation results and their discussion in Section \ref{sec:evaluation}. The paper finishes with conclusions and ideas for future research directions.

\section{Dataset}
\label{sec:datasets}
The dataset for the Sentence-Level Direct Assessment shared task is composed of data extracted from Wikipedia for six language pairs, consisting of high-resource languages English-German (En-De) and English-Chinese (En-Zh), medium-resource languages Romanian-English (Ro-En) and Estonian-English (Et-En), and low-resource languages Sinhala-English (Si-En) and Nepalese-English (Ne-En), as well as a a Russian-English (Ru-En) dataset which combines articles from Wikipedia and Reddit \cite{specia2020findings}. Each language pair has 7,000 sentence pairs in the training set, 1,000 sentence pairs in the development set and another 1,000 sentence pairs in the testing set. Each translation was rated with a score between 0 and 100 according to the perceived translation quality by at least three translators \cite{fomicheva2020unsupervised}. The DA scores were standardised using the z-score. The quality estimation systems have to predict the mean DA z-scores of the test sentence pairs \cite{specia2020findings}. 

\section{Methodology}
\label{sec:methodology}

This section presents the methodology used to develop our quality estimation methods. Our methodology is based on \TransQuest our recently introduced QE framework \cite{transquest:2020}. 
We first briefly describe the neural network architectures TransQuest proposed, followed by the training details. More details about the framework can be found in \cite{transquest:2020}.

\subsection{Neural Network Architectures}
\label{sebsec:archi}

The \textit{\TransQuest} framework that is used to implement the two architectures described here relies on the XLM-R transformer model \cite{conneau2019unsupervised} to derive the representations of the input sentences \cite{transquest:2020}. The XLM-R transformer model takes a sequence of no more than 512 tokens as input, and outputs the representation of the sequence. The first token of the sequence is always \textsc{[CLS]} which contains the special embedding to represent the whole sequence, followed by embeddings acquired for each word in the sequence. As shown below, proposed neural network architectures of TransQuest can utilise both the embedding for the \textsc{[CLS]} token and the embeddings generated for each word \cite{transquest:2020}. The output of the transformer (or transformers for \textbf{Siamese\TransQuest} described below), is fed into a simple output layer which is used to estimate the quality of translation. The way the XLM-R transformer is used and the output layer are different in the two instantiations of the framework. We describe each of them below. The fact that TransQuest does not rely on a complex output layer makes training its architectures much less computationally intensive than alternative solutions. The \textit{\TransQuest} framework is open-source, which means researchers can easily propose alternative architectures to the ones TransQuest presents \cite{transquest:2020}.


Both neural network architectures presented below use the pre-trained XLM-R models released by HuggingFace's model repository \cite{Wolf2019HuggingFacesTS}. There are two versions of the pre-trained XLM-R models named XLM-R-base and XLM-R-large. Both of these XLM-R models cover 104 languages \cite{conneau2019unsupervised}, potentially making it very useful to estimate the translation quality for a large number of language pairs.

\begin{figure*}[ht]

  \begin{subfigure}[b]{6cm}
    \centering\includegraphics[width=5cm]{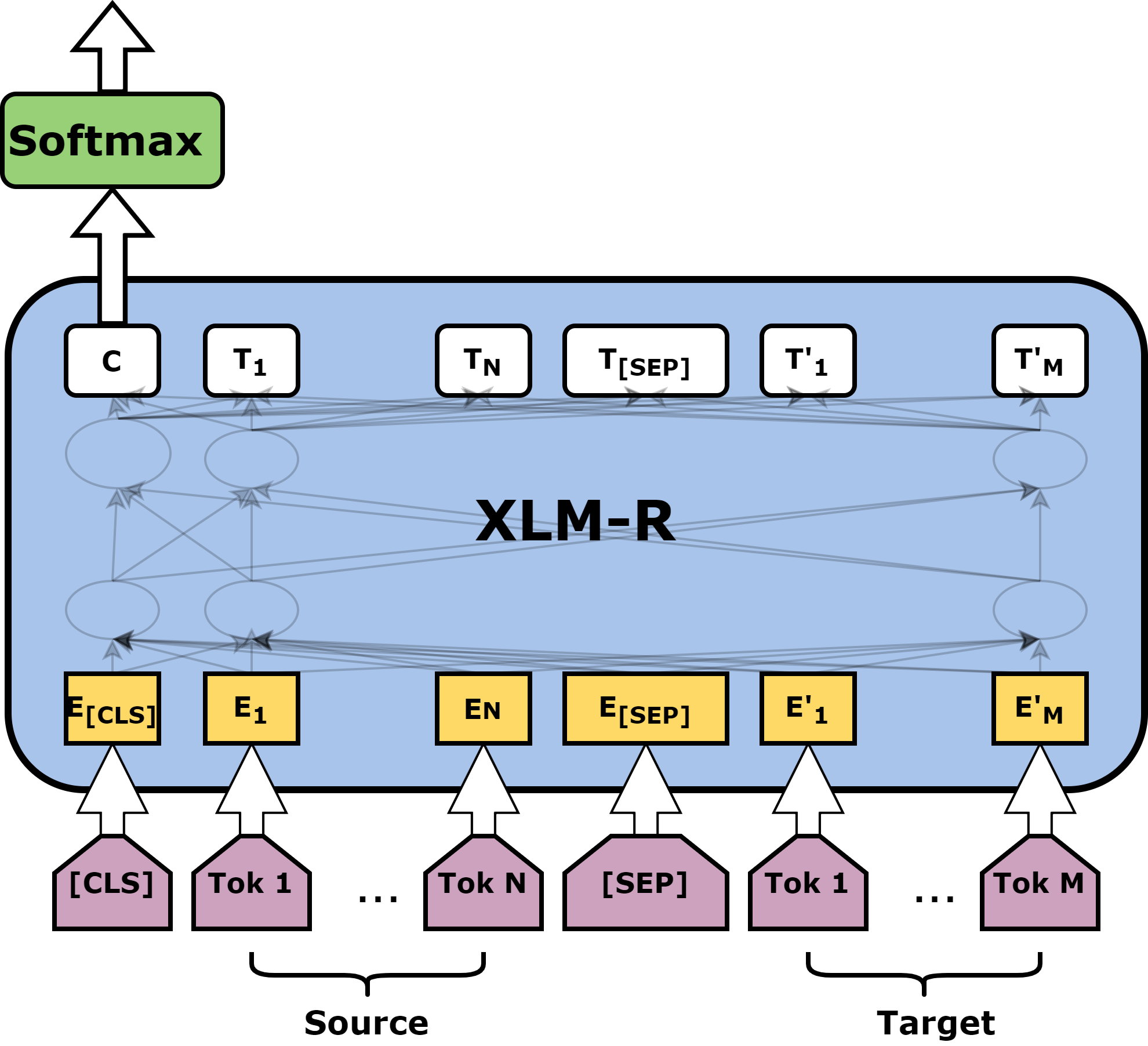}
    \caption{\textit{M\TransQuest} architecture}
    \label{fig:transquest_architecture}
  \end{subfigure}
  \begin{subfigure}[b]{6cm}
    \centering\includegraphics[width=10cm]{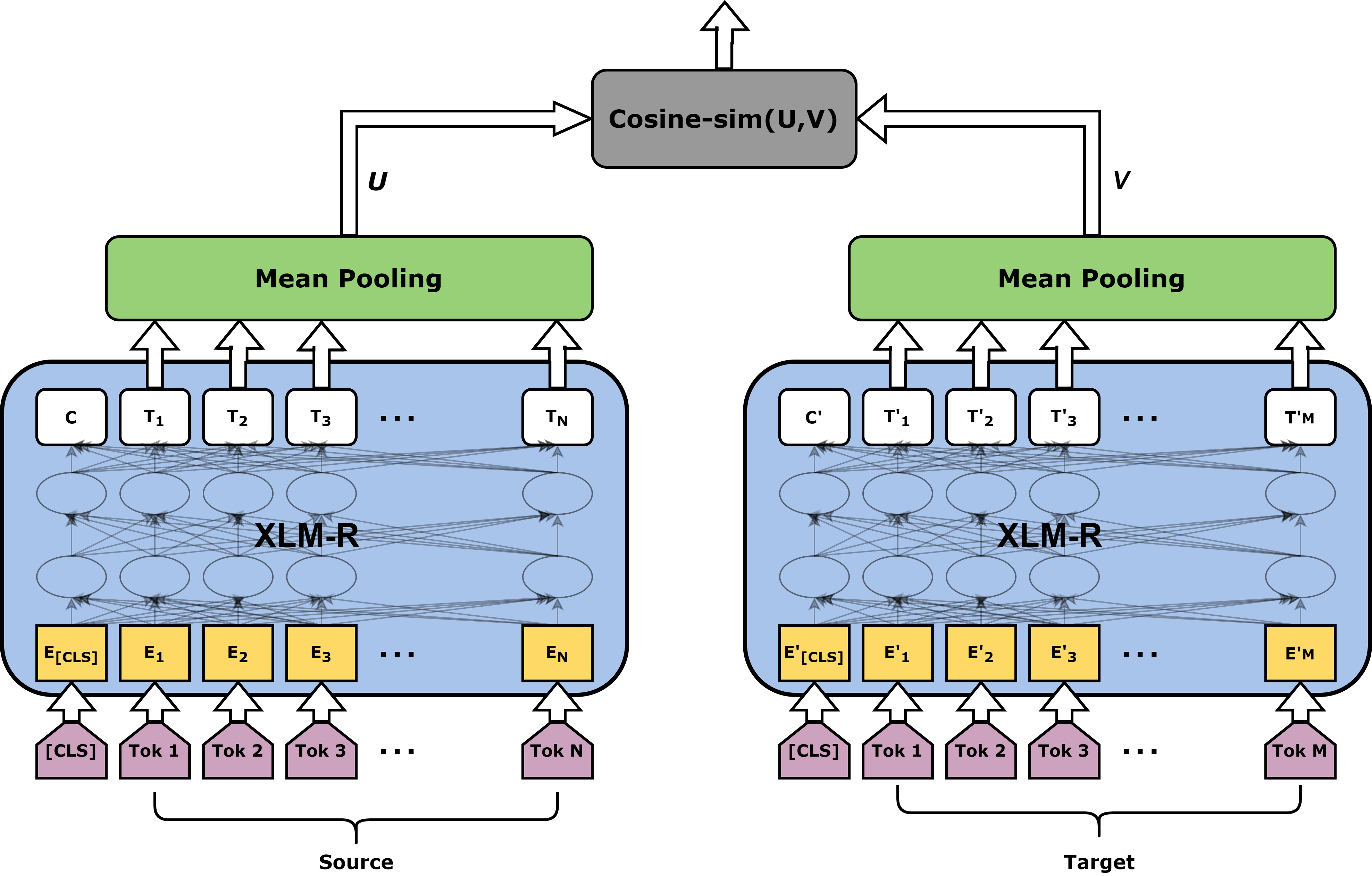}
    \caption{\textit{S\TransQuest} Architecture}
    \label{fig:siamese_transquest_architecture}
  \end{subfigure}
\caption{Two architectures of the \textit{\TransQuest} framework.}
\label{fig:architectures}
\end{figure*}

\textit{\TransQuest} implements two different neural network architectures \cite{transquest:2020} to perform sentence-level translation quality estimation as described below. The architectures are presented in Figure \ref{fig:architectures}.

\begin{enumerate}
  \item \textbf{Mono\TransQuest} (\textbf{M\TransQuest}): 
  The first architecture proposed uses a single XLM-R transformer model and is shown in Figure \ref{fig:transquest_architecture}. The input of this model is a concatenation of the original sentence and its translation, separated by the \textsc{[SEP]} token. TransQuest proposes three pooling strategies for the output of the transformer model: using the output of the \textsc{[CLS]} token (\texttt{CLS}-strategy); computing the mean of all output vectors of the input words (\texttt{MEAN}-strategy); and computing a max-over-time of the output vectors of the input words (\texttt{MAX}-strategy) \cite{transquest:2020}. The output of the pooling strategy is used as the input of a softmax layer that predicts the quality score of the translation. TransQuest used mean-squared-error loss as the objective function \cite{transquest:2020}. Similar to \citet{transquest:2020}, the early experiments we carried out demonstrated that the \texttt{CLS}-strategy leads to better results than the other two strategies for this architecture. Therefore,  we used
  the embedding of the \textsc{[CLS]} token as the input of a softmax layer.

  \item \textbf{Siamese\TransQuest} (\textbf{S\TransQuest}): The second approach proposed in TransQuest relies on the Siamese architecture depicted in Figure \ref{fig:siamese_transquest_architecture} which has shown promising results in monolingual semantic textual similarity tasks \cite{reimers-gurevych-2019-sentence,ranasinghe-etal-2019-semantic}. For this, we fed the original text and the translation into two separate XLM-R transformer models. Similarly to the previous architecture, we experimented with the same three pooling strategies for the outputs of the transformer models \cite{transquest:2020}. TransQuest then calculates the cosine similarity between the two outputs of the pooling strategy. TransQuest used mean-squared-error loss as the objective function. Similar to \citet{transquest:2020} in the initial experiments we carried out with this architecture the \texttt{MEAN}-strategy showed better results than the other two strategies. For this reason, we used the \texttt{MEAN}-strategy for our experiments. Therefore, cosine similarity is calculated between the mean of all output vectors of the input words produced by each transformer. 
  
\end{enumerate}

 \subsection{Training Details}
 \label{subsec:training}

We used the same set of configurations suggested in \citet{transquest:2020} for all the language pairs evaluated in this paper in order to ensure consistency between all the languages. This also provides a good starting configuration for researchers who intend to use \TransQuest on a new language pair. In both architectures, we used a batch-size of eight, Adam optimiser with learning rate $2\mathrm{e}{-5}$, and a linear learning rate warm-up over 10\% of the training data. The models were trained using only training data. Furthermore, they were evaluated while training using an evaluation set that had one fifth of the rows in training data. We performed early stopping if the evaluation loss did not improve over ten evaluation rounds. All of the models were trained for three epochs. For some of the experiments, we used an Nvidia Tesla K80 GPU, whilst for others we used an Nvidia Tesla T4 GPU. This was purely based on the availability of the hardware and it was not a methodological decision.

 \subsection{Implementation Details}
 \label{subsec:implementation}
 The \TransQuest framework was implemented using Python 3.7 and PyTorch 1.5.0. To integrate the functionalities of the transformers we used the version 3.0.0 of the HuggingFace's Transformers library. The implemented framework is available on GitHub\footnote{\TransQuest GitHub repository - \url{https://github.com/tharindudr/transQuest}}.
 
 \section{Evaluation, Results and Discussion}
\label{sec:evaluation}
 This section presents the evaluation results of our architectures and the fine tuning strategies that can be used to improve the results. We first evaluate the \TransQuest framework with the default setting (Section \ref{subsec:default}). Next we evaluate an ensemble setting of \TransQuest in Section \ref{subsec:ensemble}. We finally assess the performance of \TransQuest with augmented data. We conclude the section with a discussion of the results. 
 
 The evaluation metric used was the Pearson correlation ($r$) between the predictions and the gold standard from the test set, which is the most commonly used evaluation metric in WMT quality estimation shared tasks \cite{specia-etal-2018-findings,fonseca-etal-2019-findings}. We report the Pearson correlation values that we obtained from CodaLab, the hosting platform of the WMT 2020 QE shared task. As a baseline we compare our results with the performance of OpenKiwi as reported by the task organisers \cite{specia2020findings}.
 
 \renewcommand{\arraystretch}{1.2}
\begin{table*}[t]
\begin{center}
\small
\begin{tabular}{l l  c c c c c c c} 
\toprule
& & \multicolumn{2}{c}{\bf Low-resource} & \multicolumn{3}{c}{\bf Mid-resource} & \multicolumn{2}{c}{\bf High-resource}\\\cmidrule(r){3-4}\cmidrule(lr){5-7}\cmidrule(l){8-9}
 &{\bf Method} & Si-En & Ne-En & Et-En & Ro-En & Ru-En & En-De & En-Zh\\
\midrule
\multirow{2}{*}{\bf I} & M\TransQuest & 0.6525 & 0.7914 & 0.7748 & 0.8982 & 0.7734 & 0.4669 & 0.4779 \\
& S\TransQuest & 0.5957 & 0.7081 & 0.6804 & 0.8501 & 0.7126 & 0.3992 & 0.4067 \\
\midrule
\multirow{2}{*}{\bf II} & M\TransQuest-base & 0.6412 & 0.7823 & 0.7651 & 0.8715 & 0.7593 & 0.4421 & 0.4593 \\
& S\TransQuest-base & 0.5773 & 0.6853 & 0.6692 & 0.8321 & 0.6962 & 0.3832 & 0.3975 \\
\midrule
\multirow{2}{*}{\bf III} & M\TransQuest $\otimes$ & 0.6661 & 0.8023 & 0.7876 & 0.8988 & 0.7854 & 0.4862 & 0.4853 \\
& S\TransQuest $\otimes$ & 0.6001 & 0.7132 & 0.6901 & 0.8629 & 0.7248 & 0.4096 & 0.4159 \\
\midrule
\multirow{2}{*}{\bf IV} & M\TransQuest $\otimes$ - Aug & \textbf{0.6849} & \textbf{0.8222} & \textbf{0.8240} & \textbf{0.9082} & \textbf{0.8082} & \textbf{0.5539} & \textbf{0.5373} \\
& S\TransQuest $\otimes$ - Aug & 0.6241 & 0.7354 & 0.7239 & 0.8621 & 0.7458 & 0.4457 & 0.4658 \\
\midrule
\multirow{1}{*}{\bf V} & OpenKiwi & 0.3737 & 0.3860 & 0.4770 & 0.6845 & 0.5479 & 0.1455 & 0.1902 \\
\bottomrule
\end{tabular}
\end{center}
\caption{Pearson ($r$) correlation between \textit{\TransQuest} algorithm predictions and human DA judgments. Best results for each language (any method) are marked in bold. Rows I, II, III and IV indicate the different settings of \textit{\TransQuest}, explained in Sections \ref{subsec:default}-\ref{subsec:augmentation}. OpenKiwi baseline results are in Row V.} 
\label{tab:results:direct_assesement}
\end{table*}

 \subsection{\TransQuest with Default settings}
 \label{subsec:default}
 The first evaluation we carried out was for the default configurations of the \TransQuest framework where we used the training set of each language to build a quality estimation model using XLM-R-large transformer model and we evaluated it on a test set from the same language.
 
 The results for each language with \textit{default} settings are shown in row I of Table \ref{tab:results:direct_assesement}. The results indicate that both architectures proposed in \textit{\TransQuest} outperform the baseline, OpenKiwi, in all the language pairs. From the two architectures, \textit{M\TransQuest} performs slightly better than \textit{S\TransQuest}. 
 
 As shown in Table \ref{tab:results:direct_assesement}, \textit{M\TransQuest}  gained $\approx$ 0.2-0.3 Pearson correlation boost over OpenKiwi in all the language pairs. Additionally, \textit{M\TransQuest} achieves $\approx$ 0.4 Pearson correlation boost over OpenKiwi in the low-resource language pair Ne-En.
 
 \subsection{\TransQuest with Ensemble}
  \label{subsec:ensemble}
  Transformers have been proven to provide better results when experimented with ensemble techniques \cite{xu2020improving}. In order to improve the results of \TransQuest we too followed an ensemble approach which consisted of two steps. We conducted these steps for both architectures in \TransQuest.  
  
  \begin{enumerate}
  \item We train \TransQuest using the pre-trained XLM-R-base transformer model instead of the XLM-R-large transformer model in the \TransQuest default setting. We report the results from the two architectures from this step in row II of Table \ref{tab:results:direct_assesement} as M\TransQuest-base and S\TransQuest-base. 
  \item We perform a weighted average ensemble for the output of the default setting and the output we obtained from step 1. We experimented on weights 0.8:0.2, 0.6:0.4, 0.5:0.5 on the output of the default setting and output from the step 1 respectively. Since the results we got from XLM-R-base transformer model are slightly worse than the results we got from default setting we did not consider the weight combinations that gives higher weight to XLM-R-base transformer model results. We obtained best results when we used the weights 0.8:0.2. We report the results from the two architectures from this step in row III of Table \ref{tab:results:direct_assesement} as M\TransQuest $\otimes$ and S\TransQuest $\otimes$.
 \end{enumerate}

As shown in Table \ref{tab:results:direct_assesement} both architectures in \TransQuest with ensemble setting gained $\approx$ 0.01-0.02 Pearson correlation boost over the default settings for all the language pairs.  
  
 \subsection{\TransQuest with Data Augmentation}
   \label{subsec:augmentation}
 All of the languages had 7,000 training instances that we used in the above mentioned settings in \TransQuest. To experiment how \TransQuest performs with more data, we trained \TransQuest on a data augmented setting. Alongside the training, development and testing datasets, the shared task organisers also provided the parallel sentences which were used to train the neural machine translation system in each language. In the data augmentation setting, we added the sentence pairs from that neural machine translation system training file to training dataset we used to train \TransQuest. In order to find the best setting for the data augmentation we experimented with adding 1000, 2000, 3000, up to 5000 sentence pairs randomly. Since the ensemble setting performed better than the default setting of \TransQuest, we conducted this data augmentation experiment on the ensemble setting. We assumed that the sentence pairs added from the neural machine translation system training file have maximum translation quality.
   
 Up to 2000 sentence pairs the results continued to get better. However, adding more than 2000 sentence pairs did not improve the results. We did not experiment with adding any further than 5000 sentence pairs to the training set since the timeline of the competition was tight. We were also aware that adding more sentence pairs with the maximum translation quality to the training file will make it imbalance and affect the performance of the machine learning models negatively. We report the results from the two architectures from this step in row IV of Table \ref{tab:results:direct_assesement} as M\TransQuest $\otimes$-Aug and S\TransQuest $\otimes$-Aug. 
   
This setting provided the best results for both architectures in \TransQuest for all of the language pairs. As shown in Table \ref{tab:results:direct_assesement} both architectures in \TransQuest with the data augmentation setting gained $\approx$ 0.01-0.09 Pearson correlation boost over the default settings for all the language pairs. Additionally, \textit{M\TransQuest $\otimes$-Aug} achieves $\approx$ 0.09 Pearson correlation boost over default M\TransQuest in the high-resource language pair En-De.

 \subsection{Error analysis}
 \label{sebsec:error}
 
 In an attempt to better understand the performance and limitations of \textit{\TransQuest} we carried out an error analysis on the results obtained on Romanian - English and Sinhala - English. The choice of language pairs we analysed was determined by the availability of native speakers to perform this analysis. We focused on the cases where the difference between the predicted score and expected score was the greatest. This included both cases where the predicted score was underestimated and overestimated. 
 
 Analysis of the results does not reveal very clear patterns. The largest number of errors seem to be caused by the presence of named entities in the source sentences. In some cases these entities are mishandled during the translation. The resulting sentences are usually syntactically correct, but semantically odd. Typical examples are \emph{RO: În urmă explorărilor Căpitanului James Cook, Australia și Noua Zeelandă au devenit ținte ale colonialismului britanic. (As a result of Captain James Cook's explorations, Australia and New Zealand have  become the targets of British colonialism.)} - \emph{EN: Captain James Cook, Australia and New Zealand have finally become the targets of British colonialism.} (expected: -1.2360, predicted: 0.2560) and \emph{RO: O altă problemă importantă cu care trupele Antantei au fost obligate să se confrunte a fost malaria. (Another important problem that the Triple Entente troops had to face was malaria.)} - EN: \emph{Another important problem that Antarctic troops had to face was malaria.} (expected: 0.2813, predicted: -0.9050). 
 In the opinion of the authors of this paper, it is debatable whether the expected scores for these two pairs should be so different. Both of them have obvious problems and cannot be clearly understood without reading the source. For this reason, we would expect that both of them have low scores. Instances like this also occur in the training data. As a result of this, it may be that \textit{\TransQuest} learns contradictory information, which in turn leads to errors at the testing stage.
 
 A large number of problems are caused by incomplete source sentences or input sentences with noise. For example the pair \emph{RO: thumbright250pxDrapelul cu fâșiile în poziție verticală (The flag with strips in upright position)} -  \emph{EN: ghtghtness 250pxDrapel with strips in upright position} has an expected score of 0.0595, but our method predicts -0.9786. Given that only \emph{ghtghtness 250pxDrapel} is wrong in the translation, the predicted score is far too low. In an attempt to see how much this noise influences the result, we run the system with the pair \emph{RO: Drapelul cu fâșiile în poziție verticală} -  \emph{EN: Drapel with strips in upright position}. The prediction is 0.42132, which is more in line with our expectations given that one of the words is not translated. 
 
 Similar to Ro-En, in Si-En the majority of problems seem to be caused by the presence of named entities in the source sentences. For an example in the English translation: \emph{But the disguised Shiv will help them securely establish the statue.
} (expected: 1.3618, predicted: -0.008), the correct English translation would be \emph{But the disguised Shividru will help them securely establish the statue.}. Only the named entity \emph{Shividru} is translated incorrectly, therefore the annotators have annotated the translation with a high quality. However \TransQuest fails to identify that. Similar scenarios can be found in English translations \emph{Kamala Devi Chattopadhyay spoke at this meeting, Dr. Ann.} (expected:1.3177, predicted:-0.2999) and \emph{The Warrior Falls are stone's, halting, heraldry and stonework rather than cottages. The cathedral manor is navigable places} (expected:0.1677, predicted:-0.7587). It is clear that the presence of the named entities seem to confuse the algorithm we used, hence it needs to handle named entities in a proper way.  

 \section{Conclusion}
In this paper we evaluated different settings of \textit{\TransQuest} in sentence-level direct quality assessment. We showed that ensemble results with XLM-R-base and XLM-R-large with data augmentation techniques can improve the performance of \TransQuest framework.  

The official results of the competition show that \textit{\TransQuest} won the first place in all the language pairs in Sentence-Level Direct Assessment task. \textit{\TransQuest} is the sole winner in En-Zh, Ne-En and Ru-En language pairs and the multilingual track. For the other language pairs (En-De, Ro-En, Et-En and Si-En) it shares the first place with another system, whose results are not statistically different from ours. The full results of the shared task can be seen in \citet{specia2020findings}.

In the future, we plan to experiment more with the data augmentation settings. We are interested in augmenting the training file with semantically similar sentences to the test set rather than augmenting with random sentence pairs as we did in this paper. As shown in the error analysis in Section \ref{sebsec:error} the future releases of the framework need to handle named entities properly. We also hope to implement \textit{\TransQuest} in document level quality estimation too.

\bibliography{emnlp2020}
\bibliographystyle{acl_natbib}

\end{document}